\documentclass[10pt,twocolumn,letterpaper]{article}

\usepackage[pagenumbers]{cvpr} 

\usepackage{graphicx}
\usepackage{amsmath}
\usepackage{amssymb}
\usepackage{booktabs}
\usepackage{url}            
\usepackage{amsfonts}       
\usepackage{multirow}
\usepackage{graphicx}
\usepackage{comment}
\usepackage{color, colortbl}
\usepackage[dvipsnames]{xcolor}

\usepackage[pagebackref,breaklinks,colorlinks]{hyperref}

\usepackage[capitalize]{cleveref}
\crefname{section}{Sec.}{Secs.}
\Crefname{section}{Section}{Sections}
\Crefname{table}{Table}{Tables}
\crefname{table}{Tab.}{Tabs.}

\definecolor{lgray}{rgb}{0.9,0.9,0.9}

\def\confName{CVPR}
\def\confYear{2023}

\begin{document}

\title{Rapid Deforestation and Burned Area Detection using\\ Deep Multimodal Learning on Satellite Imagery}

\author{Gabor Fodor\thanks{Corresponding author. Work in progress.} \quad Marcos V. Conde\\
\\
\textbf{H2O.ai}\\
Mountain View, CA\\
{\tt\small \{gabor.fodor, marcos.conde\}@h2o.ai}\\
}
\maketitle

\begin{abstract}

Deforestation estimation and fire detection in the Amazon forest poses a significant challenge due to the vast size of the area and the limited accessibility. However, these are crucial problems that lead to severe environmental consequences, including climate change, global warming, and biodiversity loss. To effectively address this problem, multimodal satellite imagery and remote sensing offer a promising solution for estimating deforestation and detecting wildfire in the Amazonia region. This research paper introduces a new curated dataset and a deep learning-based approach to solve these problems using convolutional neural networks (CNNs) and comprehensive data processing techniques. Our dataset includes curated images and diverse channel bands from Sentinel, Landsat, VIIRS, and MODIS satellites. We design the dataset considering different spatial and temporal resolution requirements. Our method successfully achieves high-precision deforestation estimation and burned area detection on unseen images from the region. Our code, models and dataset are open source: \url{https://github.com/h2oai/cvpr-multiearth-deforestation-segmentation}.

\end{abstract}

\section{Introduction}
\label{sec:intro}

The Amazon forest, also known as Amazonia, encompasses approximately 40\% of the remaining rainforests in the world. Unfortunately, deforestation in the Amazon is reaching dangerous levels.
In 2012, Brazil achieved an unprecedented feat among tropical countries by reducing deforestation rates in Amazonia by 84\% (4,571 km2) compared to the historical peak of 2004, when 27,772 km$^2$ of forests were clear-cut~\cite{silva2021brazilamazon}.
In August 2022, deforestation in the Amazon was 1,661 km², an amount 81\% higher than that recorded in August last year~\cite{deforestation2022}.

\begin{figure}
    \centering
    \includegraphics[trim={0.125cm 0.15cm 0.15cm 0.2cm},clip, width=0.96\linewidth]{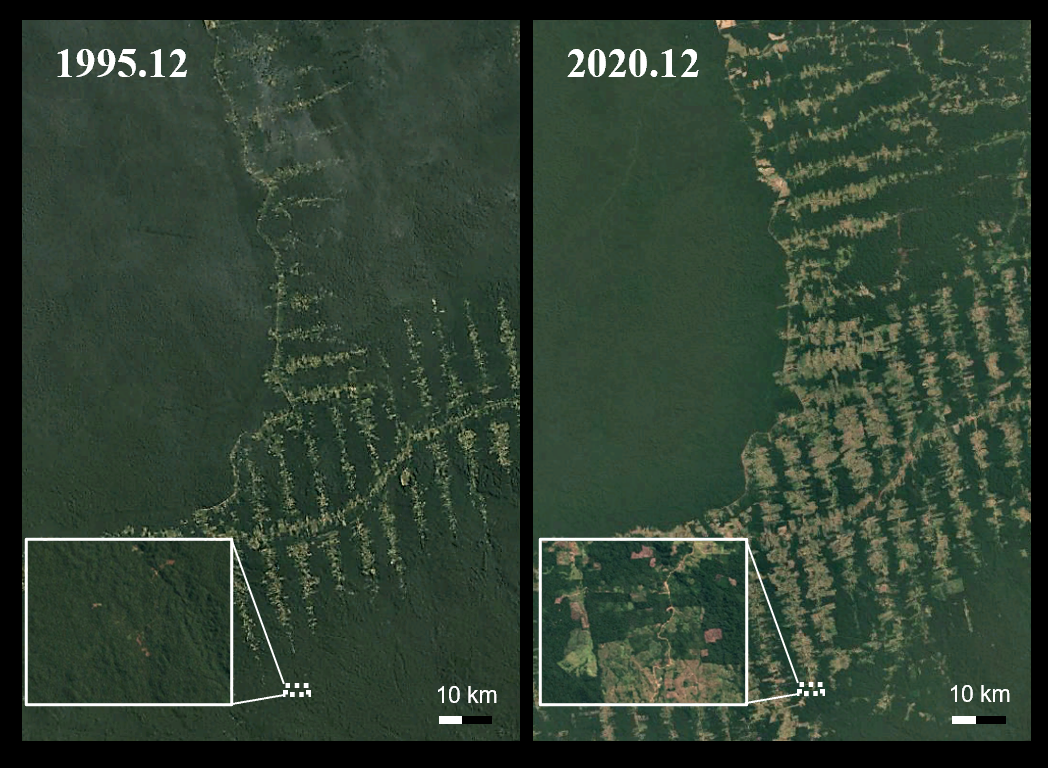}
    \caption{\textbf{Amazonia deforestation evolution}~\cite{lee2022multiearth, multiearth2022}. Image from Sentinel~\cite{sentinel}. We highlight the study area in the Amazon.}
    \label{fig:evolution}
\end{figure}

\begin{figure}[!ht]
    \centering
    \includegraphics[trim={0.15cm 0.5cm 4cm 3cm},clip, width=0.85\linewidth]{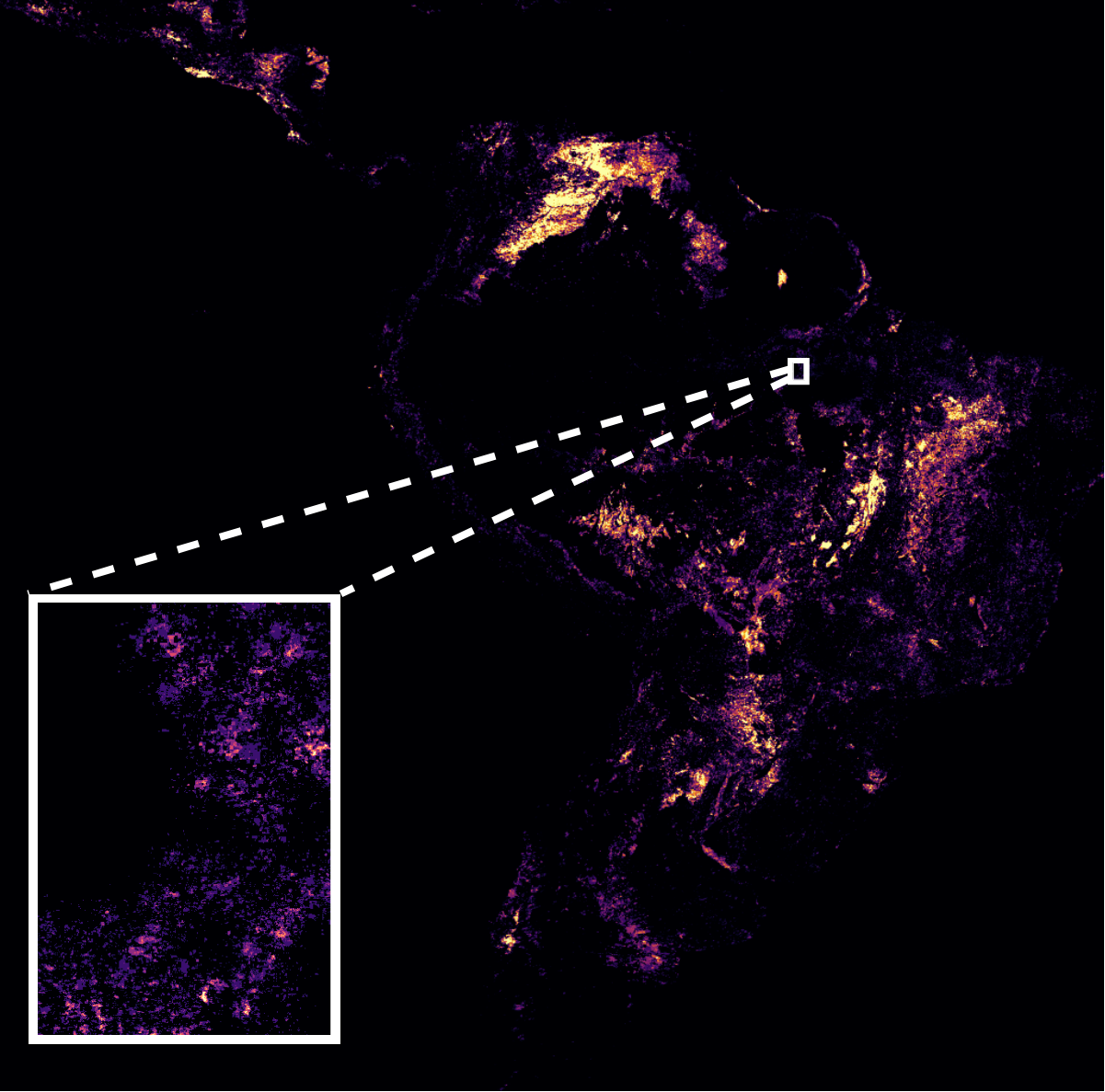}
    \caption{\textbf{South America burned areas from 2013 to 2020}.}
    \label{fig:wildfire}
\end{figure}

Deforestation estimation and burned area detection in the Amazon is very challenging due to the vast area ($\approx$ 3,000,000~km$^2$) and the limited accessibility for direct observation. Also note that this region includes territory belonging to nine nations. Despite its difficulty, scientists consider this a paramount problem since the deforestation of the Amazon leads acceleration of climate change~\cite{boulton2022deforestation, silva2021brazilamazon}.
Deforestation can be attributed to many different human-related factors. The most common -and dangerous- related factor is wildfire, which represents a major threat for societies globally causing loss of lives, infrastructures, biodiversity and ecosystems~\cite{bowman_fire_2009, pettinari_fire_2020, jones_global_2022, kondylatos_wildfire_2022}.

In this context, multimodal satellite imagery offer a powerful alternative for estimating deforestation and detecting wildfire in such regions.
Some of the most important research centers are the Brazilian National Space Research Institute (INPE), Brazilian Amazon Rainforest Monitoring Program by Satellite (PRODES) and Real-time Deforestation Detection System (DETER)~\cite{news_brazil, de2020change}.

The recent advances in machine learning (ML) allow us to improve traditional image processing using deep convolutional neural networks (CNNs). These techniques show already promising results for this task~\cite{lee2022multiearth, multiearth2022, prapas2022deep, huot_deep_2020, zhang_deep_2021, prapas_deep_2021}. As part of this effort, we can find the \textit{MultiEarth 2023}~\cite{cha2023multiearth} challenges that aim to monitor the Amazon rainforest in all weather and lighting conditions using a multimodal remote sensing dataset, which includes a time series of multispectral and synthetic aperture radar (SAR) images.
 
In this work we discuss a \textbf{deep learning framework} to support the interpretation and analysis of the Amazon rainforest (and other areas in South America) at different time and weather conditions. We tackle the two core tasks:

\begin{enumerate}
    \item Deforestation segmentation: Using images from the satellites provided by the organizers, we aim to accurately segment the deforested areas.

    \item  Fire segmentation: We suggest to use additional well maintained satellite data products to segment burned areas.
\end{enumerate}

\section{Related Work}
\label{sec:rel-work}

In the deep learning (DL) era, we can use deep convolutional neural networks (CNNs) to automatically detect and recognize patterns in images. In this work we focus on \textit{image segmentation}, the problem of identifying with a pixel-level accuracy the areas where an object (\eg person, car, building, tree, specific area) is present in an image.

Multiple works~\cite{reichstein_deep_2019, camps-valls_deep_2021} on remote sensing proposed to leverage the growing availability of open Earth Observation (EO) data and the advances in deep learning research to tackle Earth system problems.

Ban \etal introduced a method for near real-time wildfire monitoring using deep learning and images from Sentinel-1 satellite \ie Synthetic Aperture Radar (SAR)~\cite{ban2020realtimefire}.

Prapas \etal also study wildfire tracking with deep learning as a segmentation problem over time~\cite{prapas_deep_2021, prapas2022deep}.

Using such remote sensing techniques, we can collect multiple observations per day at different weather conditions. Multiple works show the potential of using Sentinel-1 to detect wildfires and capture their temporal progression, also related to deforestation tracking~\cite{rodrigues2022firo, ban2020realtimefire, lee2022multiearth, rashkovetsky2021wildfire}.

Similarly, Lee \etal used diverse channels from three satellites, Sentinel-1, Sentinel-2 and Landsat-8 to train deep neural networks for estimating deforestation in the Amazon Forest via image segmentation~\cite{lee2022multiearth}.

\paragraph{Datasets} 
Despite the importance of this problem, there is a limited amount of open-access datasets, especially for wildfire where we need variables related to the fire drivers and burned areas~\cite{prapas_deep_2021, cha2023multiearth, firecci}.
Kondylatos~\cite{kondylatos_wildfire_2022} \etal introduced a dataset of wildfire forecasting in the Eastern Mediterranean. We also find datasets that cover US areas~\cite{singla2021wildfiredb,graff2021fireml}.

\begin{table*}[h]
\centering
\resizebox{!}{0.3\columnwidth}{
\begin{tabular}{l l c c c }
    \toprule
    Satellite & Temporal Extent & Pixel Size (m) & Temporal Resolution & Bands \\
    \midrule
    Sentinel-1 & 2014 - & 10 & 6 days & VV, VH \\ 
    
    Sentinel-2 & 2018- & 10 & 5 days & B1-B9, B11, B12, QA60 \\ 
    
    Landsat 5 & 1984 - 2012 & 30 & 16 days & SR B1-B5, ST B6-B7, QA PIXEL \\
    
    Landsat 8 & 2013 - & 30 & 16 days & SR B1-B7, ST B10, QA PIXEL \\ 
    
    VNP09H1 & 2012- & 500 & 8 days & I1, I2, I3 \\
    
    VNP13A1 & 2012- & 500 & 16 days & EVI, NDVI, NIR, SWIR1, SWIR2 \\
    
    MCD15A2H & 2002- & 500 & 8 days & LAI, FPAR \\
    
    FIRMS & 2002- & 375-1000 & hours & Active fire \\
    
    \midrule
    
    Deforestation Masks & 2016 - 2021 & 10 & 1 month & Deforestation \\

    Fire CCI & 2001 - 2020 & 250 & 1 month & Burned area \\
    
    \bottomrule
\end{tabular}}
\caption{\textbf{Satellite Specifications}. We provide the time periods available for each sensor, and the selected band intervals. See also~\cite{multiearth2022}.}
\label{tab:data}
\end{table*}


\begin{figure*}[!ht]
  \centering
  \includegraphics[trim={0.125cm 1.5cm 0.15cm 0.9cm},clip, width=0.935\linewidth]{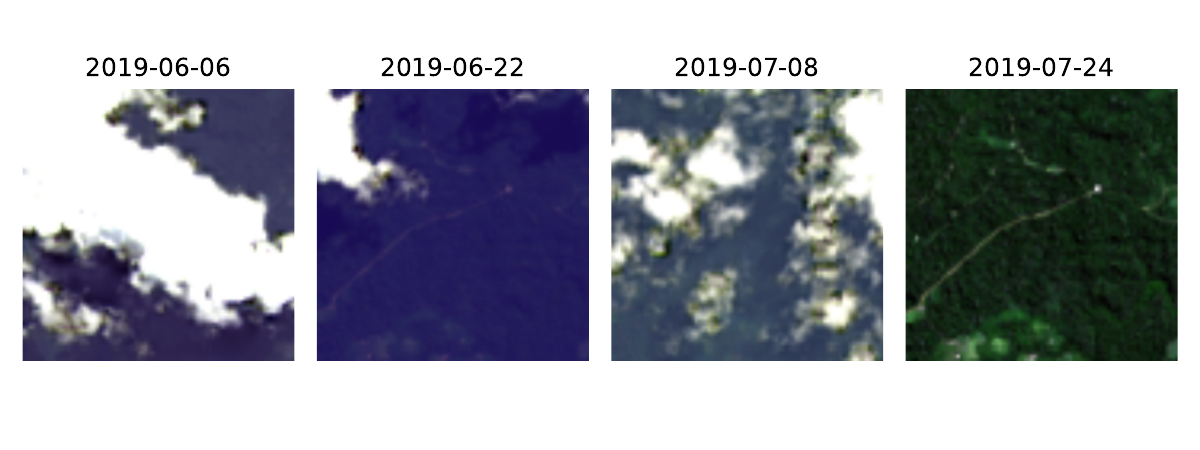}
  \caption{\textbf{Landsat-8} samples from coordinates (-54.50, -3.39) before 2019 August. This has lower resolution than Sentinel imagery.}
  \label{fig:land8}
\end{figure*}

\begin{figure*}[!ht]
  \centering
  \includegraphics[trim={0.125cm 1.5cm 0.15cm 0.9cm},clip, width=0.935\linewidth]{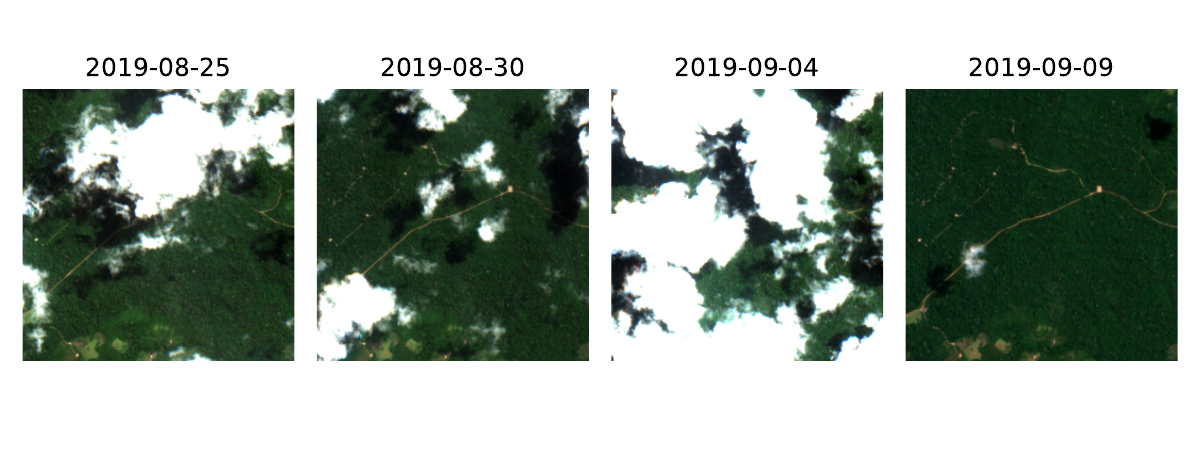}
  \caption{\textbf{Sentinel-2} samples for longitude -54.50 latitude -3.39 around 2019 August. We can observe challenging cloud occlusions.}
  \label{fig:sent2}
\end{figure*}

\begin{figure*}[!ht]
  \centering
  \includegraphics[trim={0.125cm 1.65cm 0.15cm 0.9cm},clip, width=0.935\linewidth]{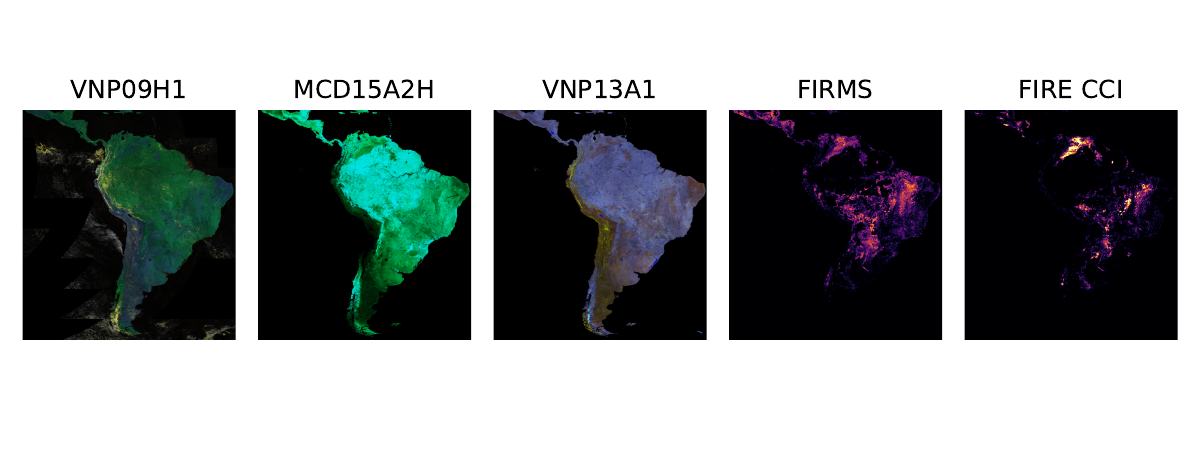}
  \caption{\textbf{Additional Satellites.} These images include surface reflectance, vegetation indices, estimated burned area maps and active fire indicators. We found this additional information very useful for predicting wildfires.}
  \label{fig:external}
\end{figure*}


\section{Proposed Dataset}
\label{sec:data}

We use as baseline the MultiEarth~\footnote{\url{https://sites.google.com/view/rainforest-challenge/multiearth-2023}} dataset~\cite{cha2023multiearth, multiearth2022}. This is a multimodal remote sensing dataset with manually annotated binary masks for deforestation, in a small region in the Amazon rainforest.

Our early experiments show that using the provided Sentinel-1, Sentinel-2, and Landsat-8 (as indicated in~\cite{cha2023multiearth}) we can successfully train neural networks to detect deforestation obtaining accurate results. However, for the particular problem of wildfire detection, we found fundamental (as we show in Section~\ref{sec:exp}) to use additional satellite data.

\paragraph{Dataset Design}

We summarize the proposed dataset in Table~\ref{tab:data}. In comparison to MultiEarth~\cite{cha2023multiearth}, our training dataset includes enhanced data products from VIIRS and MODIS satellites. The Fire Information for Resource Management System (FIRMS) was especially helpful for burned area segmentation.
Multi day high-level data products have less challenges, as each pixel represents the best possible observation during an 8-16 day period, which is selected on the basis of high observation coverage, low sensor angle, the absence of clouds or cloud shadow, and aerosol loading.

We include images captured from the different satellites in Figures~\ref{fig:land8},~\ref{fig:sent2} and ~\ref{fig:external}.

\paragraph{Sentinel and Landsat} We refer the reader to~\cite{cha2023multiearth} for more information about the data from these satellites~\cite{landsat, sentinel}.

\paragraph{FIRMS} provides near real-time (NRT) active fires and thermal anomalies from the NASA’s Moderate Resolution Imaging Spectroradiometer (MODIS)~\cite{modis} satellites, and the Visible Infrared Imaging Radiometer Suite (VIIRS) aboard S-NPP. Globally the records are available within 3 hours of satellite observation.

\paragraph{VNP09H1}~\cite{vnp09} The 8-day Visible Infrared Imaging Radiometer Suite (VIIRS) Surface Reflectance provides an estimate of land surface reflectance from the Suomi National Polar-orbiting Partnership (Suomi NPP) VIIRS sensor for three bands (I1, I2, I3) at nominal 500 meter resolution.

\paragraph{VNP13A1}~\cite{vnp13} provides vegetation indices (EVI, EVI2, NDVI) at 500 meter resolution. 
This was designed after the MODIS vegetation indices.
Along with the three vegetation layers, also includes layers for NIR reflectance -three shortwave infrared (SWIR)-; red, blue, and green reflectance.

\paragraph{MCD15A2H}~\cite{mcdsat} 8-day composite dataset with 500 meter pixel resolution. We can obtain the Leaf Area Index (LAI) and the Fraction of Photosynthetically Active Radiation (FPAR) metrics. Their algorithm chooses the best pixel available from all the acquisitions of both MODIS sensors in the 8-day period.

\begin{figure}
    \centering
    \includegraphics[trim={0.125cm 1cm 0.15cm 0.2cm},clip, width=\linewidth]{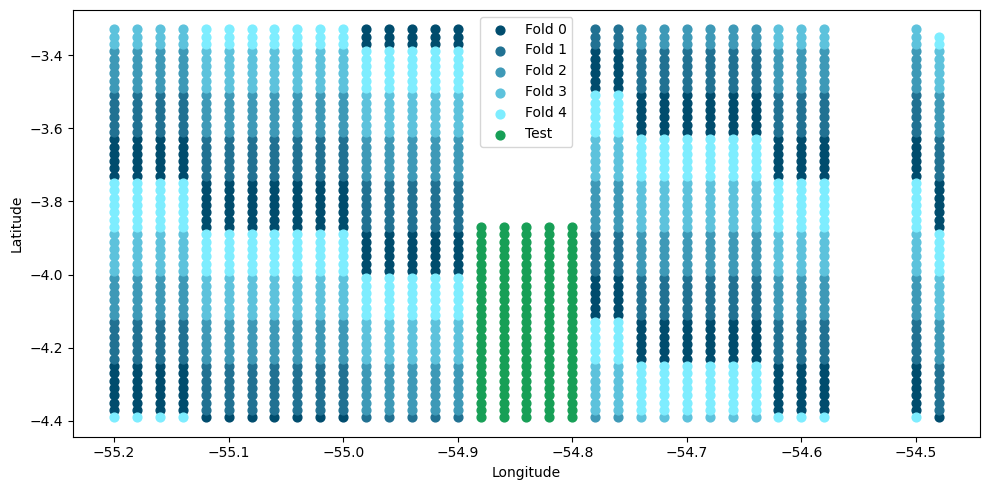}
    \caption{\textbf{Our Validation data}. Spatial 5-fold split.}
    \label{fig:folds}
\end{figure}

\begin{figure*}[!ht]
  \centering
  \includegraphics[width=\linewidth]{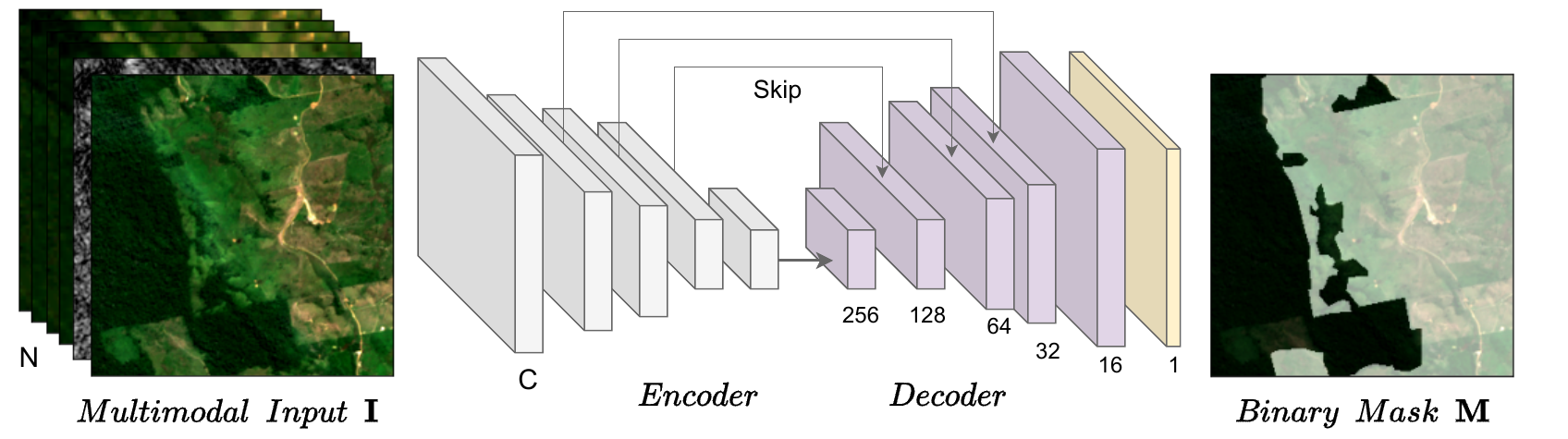}
  \caption{
    \textbf{Proposed end-to-end pipeline}. The curated multi-modal images from different satellites are stacked into a N-channel (N) representation, then this is fed into a \emph{state-of-the-art} U-Net~\cite{ronneberger2015u} segmentation model with EfficientNet~\cite{tan2019efficientnet} encoder. The output binary mask (1-channel image) is generated with a simple convolutional block (segmentation head). We show the number of channels for each decoder convolutional block. The number of input channels N varies between experiments, from 88 to 240, depending on the number of bands and recent images available.
  }
  \label{fig:pipeline}
\end{figure*}

\section{Experiments}
\label{sec:exp}

We use a common deep learning pipeline to solve deforestation and burned area segmentation, this is illustrated in Figure~\ref{fig:pipeline}. Our model requires minimal adaptation for each task \eg change the data sources and input channels.

\subsection{Deep Learning Model}
\label{sec:model}

We use a \emph{state-of-the-art} image segmentation architecture based on U-Net~\cite{ronneberger2015u} with EfficientNet-B3~\cite{tan2019efficientnet} as encoder. The model is illustrated in Figure~\ref{fig:pipeline}. We follow the standard encoder-decoder structure with skip connections and segmentation head (\ie simple convolutional layer to generate the output). For a given multi-modal input map $\mathbf{I}$, the model generates a binary mask $\mathbf{M}$ indicating deforestation (or burned area) \ie $\mathbf{M} \in [0,1]$, where 1 indicates the occurrence of the problem, and 0 indicates the area is safe.

We use this model to solve both tasks, with necessary changes in the data loaders. The training procedure is very similar, using the corresponding ground truth binary masks for each task.

\subsection{Deforestation Segmentation}

To solve this task, we use only the provided datasets from Sentinel 1-2 and Landsat 8 (as in~\cite{cha2023multiearth, lee2022multiearth}).
The Landsat 8 images were upsampled to match the 10-meter pixel resolution of the masks. 
We believe the additional vegetation indices (Enhanced Vegetation Index, Fraction of Photosynthetically Active Radiation, etc.) and related metrics would be useful, however the resolution difference is too large.
Given ideal weather conditions and continuous observations the deforestation segmentation is trivial, missing satellite data and frequent cloud occlusion makes it a challenging task. 

Instead of training separate models for each input satellites and apply post-processing later as~\cite{lee2022multiearth}, we propose training a single deep neural network on multi-channel images combining all available input dates and bands.
Since we did not have exact timestamps for the segmentation label (binary masks), we extracted all relevant bands considering a $\pm2$ months window.

\paragraph{Validation}
As the annotated binary masks were created for a small region across several years, the samples show high spatial and temporal correlation. 
To simulate the train/test separation we used spatial grid for cross validation (see Figure~\ref{fig:folds}). 
In the training region there is still a large forested area where deforestation has not happened yet; the models could learn this quickly, therefore, we downsampled this region to avoid ``easy regions" that could boost the validation score while tuning network parameters. 

\paragraph{Training}
The model (see Section~\ref{sec:model}) was trained by minimizing the Binary Cross Entropy Loss (BCE) with batch size 32, using AdamW optimizer with cosine scheduler, during 10 epochs.
We used simple non-destructive \emph{augmentations} from the dehidral group \eg horizontal and vertical flips, transpose and $90^{\circ}$~rotation. Further rotation or scaling did not help the generalization to the unseen test region. 

As the test queries had even more missing data we implemented satellite input dropout, up to half of the available imagery dates were forgotten randomly during training.

The complete implementation details can be consulted in our github project page. 

\paragraph{Results}
Each experiment takes 2-3 hours using 3 NVIDIA Quadro RTX 6000 cards. Our best single model achieves 90.4 Pixel Accuracy, 0.871 F1 and 0.792 IoU on the unseen test area (see Figure~\ref{fig:folds}). Training several similar models with small configuration changes and blending raw predictions before binarization gives small performance boost up to 90.8 Pixel Accuracy, 0.876 F1 and 0.799 IoU.

\begin{table}[]
    \centering
    \small
    \begin{tabular}{c c c c c}

        \toprule
        Rank & Team & Accuracy & F1 & IoU \\
        \midrule
        1 & FOREVER\_v2 & 91.14 & 0.89 & 0.81 \\
        \rowcolor{lgray} 2 & Ours & 90.77 & 0.88 & 0.80 \\
        3 & Ours (single model) & 90.45 & 0.87 & 0.79 \\
        4 & S.V.ccl & 88.73 & 0.84 & 0.76 \\
        5 & Star.Vision.wll & 88.43 & 0.84 & 0.75 \\
        6 & S.V & 86.90 & 0.80 & 0.72 \\
        7 & SpaceVision4Amazon & 84.14 & 0.80 & 0.69 \\
        \toprule

    \end{tabular}
    \caption{\textbf{Deforestation results} on the Multiearth 2023~\cite{cha2023multiearth} dataset. We indicate the pixel accuracy, F1-score and IoU of the predicted segmentation masks \emph{w.r.t} the ground-truth images.}
    \label{tab:deforestation-results}
\end{table}

\subsection{Fire Segmentation}

Half of the test queries were collected before 2013, we could only use Landsat 5 images from the provided datasets. The infrequent images and cloud occlusions made much more challenging to detect monthly burned areas accurately. We had a few similar binary segmentation experiments with moderate results (AUC~0.75).

The ground truth \texttt{FIRE\_CCI} dataset has much lower resolution than the provided satellite imagery; it allows us to use external datasets with higher temporal resolution. All additional datasets (\emph{w.r.t} Multiearth~\cite{cha2023multiearth}) are still maintained allowing us to train global fire detection systems.

For a larger experiment we downloaded 8 years satellite imagery covering all South America. Thanks to the lower resolution, the total dataset takes less than 100GB, allowing us to train segmentation models for full South America.

Every dataset was upsampled to match the Fire CCI~\cite{firecci} resolution and the whole region were split into 256x256 tiles keeping only the relevant regions with at least 1/5 land surface.
As we do not need manual annotations we suggest temporal validation split and use 2020 as validation set. 

Using all suggested the proposed dataset~\ref{sec:data}, from the target month and the previous three months, we are able to detect burned areas with 0.95 AUC for South America.

\section{Conclusion}

We present a deep learning model for wildfire and deforestation detection. Our  deforestation model achieves great performance combining all provided data sources.
The proposed pipeline is easy to generalize and adapt for other satellite imagery  tasks.
Additionally, we propose new datasets for rapid fire detection at global scale. Early fire segmentation experiments show promising results by reaching 0.95 pixel-wise AUC for the South America region. Using the studied datasets, it is feasible to build global fire hazard prediction models.

{
\bibliographystyle{ieee_fullname}
\bibliography{references.bib}
}

\end{document}